\documentclass{article}

    \PassOptionsToPackage{numbers, compress}{natbib}

     \usepackage[final]{neurips_2019}

\usepackage[utf8]{inputenc} 
\usepackage[T1]{fontenc}    
\usepackage{hyperref}       
\usepackage{url}            
\usepackage{booktabs}       
\usepackage{amsfonts}       
\usepackage{nicefrac}       
\usepackage{microtype}      
\usepackage{natbib}
\usepackage{graphicx}
\usepackage{afterpage}
\usepackage{amsmath}
\usepackage{caption}
\usepackage{subcaption}
\usepackage{multicol}
\usepackage{blindtext}

\title{GPoeT-2: A GPT-2 Based Poem Generator}

\author{%
  Kai-Ling Lo\\
  School of Computer Science\\
  Carnegie Mellon University \\
  Pittsburgh, PA 15213 \\
  \texttt{kailingl@cs.cmu.edu} \\
   \And
   Rami Ariss \\
   Civil and Environmental Engineering \\
   Carnegie Mellon University \\
   Pittsburgh, PA 15213 \\
   \texttt{rariss@andrew.cmu.edu} \\
   \AND
   Philipp Kurz \\
   School of Computer Science\\
   Carnegie Mellon University \\
   Pittsburgh, PA 15213 \\
   \texttt{kurz@cmu.edu} \\
}

\begin{document}

\maketitle

\begin{abstract}
    This project aims to produce the next volume of machine-generated poetry, a complex art form that can be structured and unstructured, and carries depth in the meaning between the lines. GPoeT-2 is based on fine-tuning a state of the art natural language model (i.e. GPT-2) to generate limericks, typically humorous structured poems consisting of five lines with a \textit{AABBA} rhyming scheme. With a two-stage generation system utilizing both forward and reverse language modeling, GPoeT-2 is capable of freely generating limericks in diverse topics while following the rhyming structure \textbf{without any seed phrase or \textit{a posteriori} constraints}.Based on the automated generation process, we explore a wide variety of evaluation metrics to quantify "good poetry," including syntactical correctness, lexical diversity, and subject continuity. Finally, we present a collection of 94 categorized limericks that rank highly on the explored "good poetry" metrics to provoke human creativity. \\~\\
    This project is open-sourced and the codebase is available \href{https://github.com/coderalo/11785-automatic-poetry-generation}{here}.
\end{abstract}

\section{Introduction}
Creative artificial intelligence (AI) can push the boundary of human creativity by generating new content that provokes the question: "What makes good art?" To explore creative AI, we focus on poetry, specifically limericks. Limericks are structured poems, typically humorous, consisting of five lines with an \textit{AABBA} rhyming scheme. The structured nature of limericks and complexity of short-form storytelling and humor make it well-suited to the task of creative AI exploration. 

In order to automatically generate poetry, it is critical to first define what "good poetry" is such that it can be quantified and learned by a machine. As interpretation of "good poetry" is highly subjective, we first acknowledge that metric selection in in of itself is also a subjective pursuit. While measuring the quality of AI generated poems may be done with human evaluation via a Turing test (i.e. characterizing the quality of a generated poem based on how likely it is for humans to mistake it for a human-generated poem), such evaluations are difficult to scale with consistency and provide little insight into why a poem may actually be "good." We instead evaluate poems with various scoring functions for metrics such as lexical diversity and subject continuity that quantify the qualities of a poem. This allows us to programatically define "good poetry" and embed evaluation, ranking, and selection of generated poems into GPoeT-2. Having such scalable, quantifiable metrics also allows the creation of custom loss functions for fine-tuning language models that improve the overall model performance for poem generation tasks.

The novelty of exploring evaluation metrics for good poetry to refine and select from infinitely producible AI poems can result in an automated generation process whose output can be novel poems that provoke human thought and creativity. Having indistinguishable AI-generated poems from human-written poems is a secondary objective to that of producing "good poetry" that is interpretable, thought-provoking, and creative.



\section{Related Works}
\label{lit_review}

In general, poetry generation falls under the umbrella of language generation tasks. We rely on vast amounts of pre-existing natural language models (LMs) tuned to poetry structures with constrained generation and discriminative evaluation.

\subsection{Language Generation}
Neural network based models, especially recurrent neural networks (RNN) and transformers \cite{VaswaniAttentionNeed}, can be traced back to the early 2000s \cite{Bengio2003AModel} and are nowadays commonly used for both unconditional and conditional language generation. While RNN-based models were dominant in the first half of the 2010s \cite{SeeGetNetworks, Luong2015EffectiveTranslation, SundermeyerLSTMModeling}, transformers quickly became the new paradigm for language generation after their introduction in 2017.

Recently, large-scale unsupervised pre-training \cite{Devlin2018BERT:Understanding} finds success in almost all natural language processing (NLP) tasks, including both unconditional and conditional language generation. Built upon traditional left-to-right language modeling, the GPT \cite{OpenaiImprovingPre-Training} series (especially GPT-2 \cite{RadfordLanguageLearners} and GPT-3 \cite{Brown2020LanguageLearners}) are applied to a wide variety of language generation tasks with superior performance compared to state-of-the-art methods, while pre-trained sequence-to-sequence models \cite{Wang2019BERTModel, Zhang2020PEGASUS:Summarization} and masked language models \cite{Lewis2019BART:Comprehension, Rothe2020LeveragingTasks} have proven to be capable of generating high-quality text.

\subsection{Language Representation}
Language representation maps tokenized words to real-valued vectors in an embedding space where similarities and word associations can be quantified using distance metrics. A classic example is that a well-trained word embeddings can capture analogies such as "King - Man + Woman = Queen." 

While the early approaches of Word2Vec and GloVe \cite{MikolovEfficientSpace, PenningtonGloVe:Representation} mainly focus on capturing word features by only implicitly using the context during training phase, recent works equipped with large-scale language-model based pre-training (e.g. ELMo, BERT) \cite{PetersDeepRepresentations, Devlin2018BERT:Understanding} were proven to be able to better capture context information and generate more generalized language representations.

Similar to word embeddings, lexical databases and ontologies also aim to interconnect words and concepts in a semantically meaningful way. WordNet \cite{Miller1995WordNet:English} is a manually constructed lexical database for the English language that relates words to each other in a hierarchical network structure. By grouping words into sets of synonyms and interrelating words with similar meaning, the database can be leveraged to extract semantic relationship information of given word pairs.

\subsection{Poetry Generation}
Unlike typical language generation tasks, such as translation or summarization, traditional poems (e.g. haiku and limericks) adhere to a pre-defined verse structure and a set of characteristics that introduce non-trivial difficulty for automatic generation. While some early attempts relied on rule-based and template-based methods \cite{NetzerGaikuNorms, Wu2009NewSystem} or statistical language models \cite{He2021GeneratingModels}, most of the recent works are based on neural networks, including attentional RNN \cite{Zhang2014ChineseNetworks, HopkinsAutomaticallyNetworks, JhamtaniLearningAdversaries} and transformers \cite{BenaIntroducingGeneration}.

It is worth mentioning that a recent work \cite{Luong2015EffectiveTranslation} built upon template-based method and beam search outperformed state-of-the-art neural network-based systems, indicating that there is still lots of room for improvement for deep learning-based poetry generation. Another recent work \cite{SeeGetNetworks} explored the possibility of generating poems with only training on prosaic text, which sheds light on how neural networks, potentially pre-trained on a large corpus, can be used for generating high-quality poems through enforcing \textit{a priori} constraints.

\section{Methods}
\label{gen_inst}

\begin{figure}
  \centering
  \includegraphics[width=.9\textwidth]{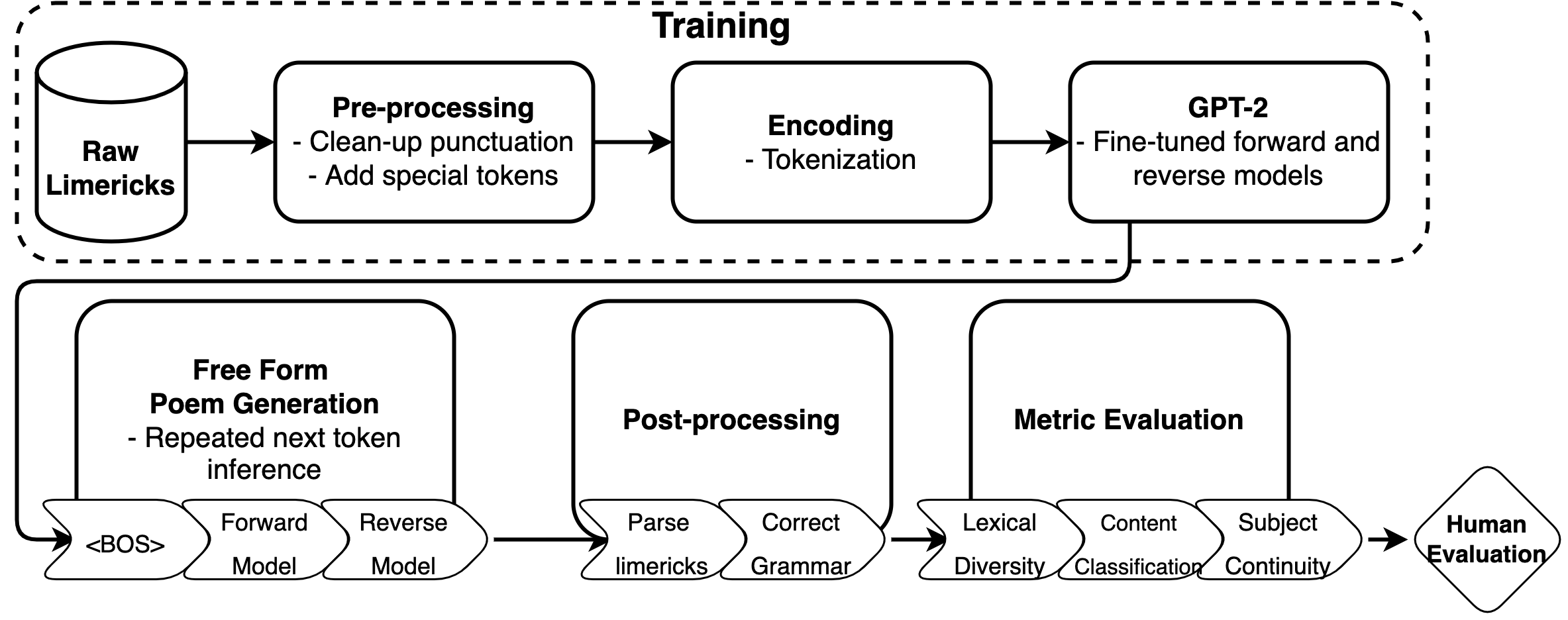}
  \caption{Model training and automatic poem generation process for GPoeT-2.}
  \label{fig:pipeline}
\end{figure}

Our system fine-tunes state-of-the-art language models to generate limerick structures through a two-stage generation process that utilizes both forward and reverse language modeling. The full pipeline is illustrated in Figure \ref{fig:pipeline}.

\subsection{Data Pre-processing and Encoding}
Before feeding the limericks as input to the model, we introduce the following pre-processing and encoding steps:
\begin{enumerate}
    \item Insertion of the additional special tokens \texttt{<BOS>} at the beginning of a poem and \texttt{<LINE>} between the lines to help the model better identify the structure of the poems.
    \item Tokenization of poems that follows the GPT-2 format.
    \item For reverse language modeling, reverse the order of tokens of each line in the limericks, while the order of the lines is still maintained.
\end{enumerate}
The first and third steps are crucial to high-quality fine-tuning of both the forward and reverse language model, which result in both naturally coherent structure and good rhyming without additional constraint during generation. 

\subsection{Model: Fine-tuning GPT-2 for Poetry Generation}
 \href{https://openai.com/blog/better-language-models/}{GPT-2} \cite{RadfordLanguageLearners} is a generative language model built based on the generator part of the transformer architecture \cite{VaswaniAttentionNeed}, which generates text samples from an arbitrary language input (see Figure \ref{fig:gpt-2_arch}). By fine-tuning it on poetry datasets, we can utilize it for both conditional and unconditional poetry generation. 

\afterpage{%
\begin{figure}
  \centering
  \includegraphics[width=.7\textwidth]{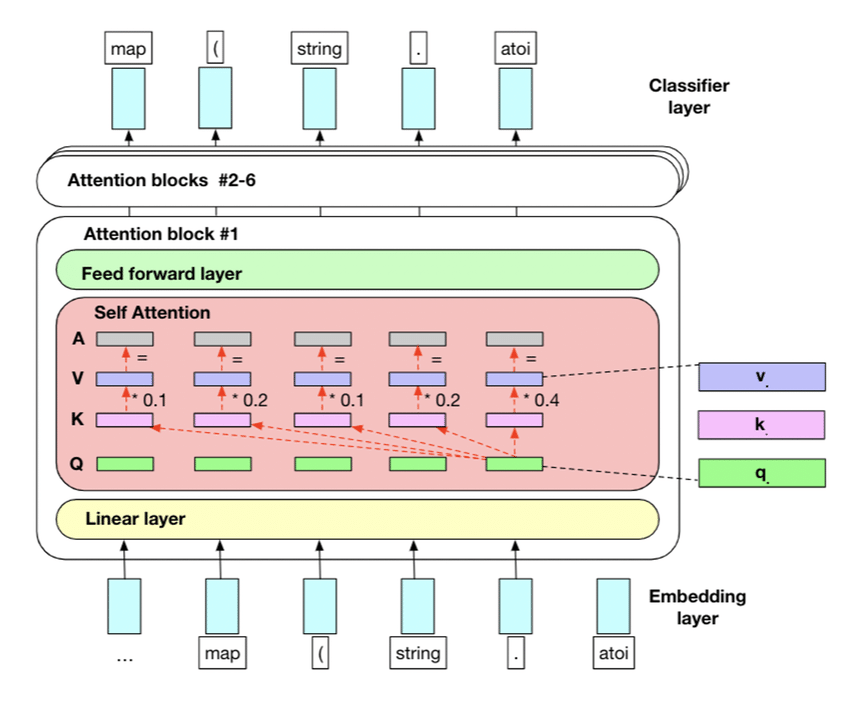}
  \caption{Example of GPT-2 transformer architecture \cite{AyeLearningDatasets}}
  \label{fig:gpt-2_arch}
\end{figure}
}
 
 GPT-2 is trained and fine-tuned to minimize the negative log-likelihood loss over the sequence:
 $$
 L(\theta) = -\sum_{t=1}^T\log \mathbb{P}(w_t \mid w_0, w_1, w_2, \ldots, w_{t-1}; \theta),
 $$
 with model parameters $\theta$, and tokens for each input sequence (i.e. limerick) $w_0, w_1, w_2, \ldots, w_T$. We fine-tune the GPT-2 model for 20 epochs using the default settings in the Transformers \cite{WolfTransformers:Processing} package using a limerick dataset (see section \ref{dataset_description}).

To pick the epoch checkpoint to use for limerick generation, we calculate the perplexity on the held-out validation set and choose the checkpoint with the lowest perplexity. Perplexity indicates how confused the model is when it tried to generate the subset of the corpus, where lower perplexity indicates a more accurate language model. More formally, perplexity is the "multiplicative inverse of the probability assigned to the test set by the language model, normalized by the number of words in the test set."

 \subsubsection{Forward LM Fine-tuning}
 The forward language model utilizes the standard order (left-to-right) of tokens within each limerick as the fine-tuning corpus for GPT-2. We notice subpar performance with this fine-tuned language model; while it does generally capture the subject continuity, it does not learn to generate the \textit{AABBA} rhyming structure of limericks.
 
 \subsubsection{Reverse LM Fine-tuning}
 By fine-tuning the model with a corpus of reverse order (right-to-left) tokens within each line of each limerick while retaining the order of the lines within the limerick, the model successfully learns to generate the \textit{AABBA} rhyming structure. However, the base GPT-2 model is trained as a forward language model, and this prior results in the failure of the reverse LM to generate high-quality limericks without a given seed line or phrase at the beginning of generation.

\subsection{Two-Stage Free Form Limerick Generation}
To generate free form limericks--that is, limericks generated without a given seed line or phrase--we introduce a two-stage generation technique to utilize the best from the forward and reverse language models. Two-stage generation uses the forward language model to generate a limerick's first line with high quality and diversity, then uses the reverse language model to generate the rest of the four lines given the first line generated by forward LM.

\subsection{Poem Evaluation}
For GPoeT-2, we subjectively select and evaluate metrics that quantify the idea that "good poetry" should be:

\begin{itemize}
    \item syntactically correct (i.e. correct grammar, valid words, valid punctuation and capitalization)
    \item lexically diverse (i.e. not be overly repetitious and have a high ratio of unique poems to overall tokens in a poems)
    \item consistent in subject and topic (i.e. nouns/subjects should be related and poems should have an identifiable topic(s))
\end{itemize}

\subsubsection{Post-processing}
As the generated limericks are produced with repeated next token inference with a maximum length, we first post-process a generated batch of poems into parsed limericks. We then filter out syntactically incorrect poems and automatically correct simple mistakes.

\textbf{Syntactical Correctness}
Next token inference makes it possible for GPoeT-2 to generate incorrect words, grammar, and invalid punctuation. A simple evaluation metric to capture and penalize such limericks is evaluating for syntactical correctness. Using an open-source spell and grammar checker, we filter out limericks that contain incorrect words and language and automatically correct simple grammar errors, such as lower-case "I" pronouns, for the remaining poems \cite{LanguageToolChecker}. This metric essentially quantifies the assumption that "good poetry" should be syntactically correct.

\subsubsection{Metric Evaluation}
Given post-processed limericks, we calculate metrics for "good poetry", such as lexical diversity, to rank and select individual limericks and rate the quality of the automatic poetry generation model. 

\textbf{Lexical Diversity} For Lexical diversity, the ordinary Type-Token Ratio (TTR) is calculated for each limerick. TTR is the ratio of number of unique tokens (V) to the number of total tokens (N) in a limerick:
\begin{equation*}
    TTR = \frac{V}{N}
\end{equation*}

The larger the ratio, the more likely we expect the limerick to be of high quality with variations in the rhymed words at the end of each line in the \textit{AABBA} rhyming scheme, as well as using a more diverse vocabulary overall.

\textbf{Subject Continuity}
We define good poems as those that tell a coherent story, with individual lines building onto each other logically to provide a meaningful flow of information. Automatically generated poems, however, are susceptible to erratically switching topics from line to line and using undefined pronouns and dissociated nouns, resulting in inconsistent storytelling throughout the poem. Quantifying subject continuity therefore becomes important for ranking poems and improving model performance.

\textit{\textbf{BERT-based Embedding Distances}}
To derive a measure for semantic similarity across multiple words, we propose the use of word embeddings, which project words into a high-dimensional numerical space based on their meaning and context. While most words in a given poem provide at least some degree of semantic information, we argue that just the nouns themselves used throughout the poems can serve as a good proxy for the poem's actual subject while avoiding the noise associated with the rest of the poems context. Hence, assuming that distances in the word embedding space correlate with semantic similarity of a pair of words, we propose to quantify subject continuity throughout a poem as the average noun centroid distance in the embedding space. Given a word embedding function $f(w) \in \mathbb{R}^k$ that projects a given word $w$ into $k$-dimensional space, and a sequence of nouns $S_w = \{w_1, w_2,...,w_n\}$ from the poem, we suggest to first calculate the embedding centroid $f_C(S_w) \in \mathbb{R}$ as the dimension-wise average embedding:
\begin{equation*}
    f_C(S_w) = \frac{1}{n} \sum\limits_{i = 1}^{n} f(w_i)
\end{equation*}

Then, the euclidean distance $D(w_i)$ for a given noun $w_i$ to the centroid can be calculated:
\begin{equation*}
    \forall i \in \{1,...,n\}: \; D(w_i) = \lVert f_C(S_w) - f(w_i)\rVert^2
\end{equation*}

We can then calculate the mean and standard deviation of all nouns' distances to the centroid. Intuitively, the mean distance indicates how spread out the nouns are: the higher the mean, the further apart the nouns are from the average poem subject in embedding space. The standard deviation indicates if there are any outlier nouns with respect to subject continuity: if the variance is low, it intuitively suggests that all nouns are equally as close or far from the average poem subject, whereas a high variance implies that some nouns are close to the subject and others are not. The latter case might suggest multiple subjects in the poem.

\textit{\textbf{WordNet-based Similarity Metric}}
homogoneous not continuous subjects

The approach using BERT centroid distances relies on the assumption that subject continuity is achieved by a poem with semantically similar words. It incentives subject homogeneity rather than continuity. While embedding distances actually capture information on relationships and analogies between words, a metric for subject continuity is difficult to derive.

To address this problem, we propose an alternative to calculating noun similarity through distances in the embedding space: leveraging graph-based ontologies and the shortest path distance between words in the graph. A common ontology used in NLP tasks is WordNet, which is a lexical database that groups words in a hierarchical structure from least specific to most specific, and interrelates words that are synonyms to each other. Using this graphical representation of words and their relations, the similarity can be defined as the shortest path in the graph, normalized to a value in the range from 0 to 1. To compute a score for subject continuity using WordNet, we calculate the average pairwise distance between all nouns in the poem. A small pairwise distance implies that the nouns are relatively close to each other in terms of their semantic meaning, whereas large pairwise distances imply unrelated semantic meaning.

\textit{\textbf{Content Classification}}
Text classification--assigning labels to textual units--is a classic and well studied problem in NLP \cite{Minaee2020DeepReview}. Content classification tries to identify the theme or topic of a text, assigning it a category. Automated content classification tools are readily available, and we query Google's Natural Language API Content Classification service to produce topic(s) and their associated confidence metrics for each limerick \cite{ContentCloud}. The content classification tool classifies text into one or more categories from a predefined list of 700+ predefined categories, with details and subcategories such as "/Health/Health Conditions/Diabetes" and "/Science/Ecology \& Environment". We utilize the proprietary tool as a black box as no publicly available information is available detailing the underlying model. 

To utilize the classified categories as a proxy for subject continuity, we assume a correlation between the maximum confidence scores of a poems assigned categories and the overall topic continuity of the poem. It is important to note that idiosyncratic language, which is prevalent in our generated limericks due to our training dataset, results in unclassified poems when using Google's Content Classification service. While we filter out unclassified poems in our ranking and selection of generated poems, we underscore that unclassified poems do not necessarily signify poor subject continuity. Rather, unclassified poems might also be a product of idiosyncratic language prevalent in the poem. For either reason, we choose to filter out such poems as idiosyncratic language is unlikely to be universally understood and connected with by a general audience, therefore making it less likely for such a poem to be "good."


\section{Results}
We generate various sets of limericks using the different language models, measure their quality after post-processing using lexical diversity and subject continuity metrics, and manually evaluate a subset of the poems to produce a collection of 94 AI limericks.


\subsection{Dataset Description}
\label{dataset_description}


\href{http://www.oedilf.com/db/Lim.php}{The Omnificent English Dictionary In Limerick Form (OEDILF)} provides a substantially large database of 113,722 approved limericks accessible to the public. OEDILF is compiling an online dictionary to "write at least one limerick for each meaning of each and every word in the English language." Their definition of a "good limerick" as one that clearly defines the the meaning of the English word in a humorous or interesting way may potentially bias our poetry generation.  
An example of a limerick in this dataset format defining the word \textbf{benthic} is as follows:

\begin{quote}
    cap'n jack was washed over the side.\\
    his crew searched but found not hair nor hide.\\
    no longer the helm,\\
    but the deep \textbf{benthic} realm,\\
    is where jack will forever reside.\\
\end{quote}

\subsection{Quantitative Results}

\subsubsection{Limerick Rhyming Structure Performance}

For both the standard order and reverse order language models, we fine-tune two GPT-2 checkpoints, which have 176M (small) and 431M (medium) parameters respectively, on the OEDILF limerick dataset. We then calculate the perplexity on a held-out validation set of limericks. During training, the perplexity with respect to limerick generation significantly decreases from the original GPT-2 perplexity, which is around 60-70 for forward language model (see Table \ref{perplexity-table}).

\begin{table}
  \caption{Perplexity on the held-out validation set of limericks}
  \label{perplexity-table}
  \centering
  \begin{tabular}{ll}
    \toprule
    Model & Perplexity $\downarrow$ \\
    \midrule
    GPT-2 (small, forward LM) & 18.3541 \\
    GPT-2 (medium, forward LM) & 16.5147 \\
    GPT-2 (small, reverse LM) & 20.5904 \\
    GPT-2 (medium, reverse LM) & 25.3487 \\
    \bottomrule
  \end{tabular}
\end{table}

While the perplexity of forward LMs are lower than the perplexity of reverse LMs, we observe that only reverse LMs successfully learn to generate the \textit{AABBA} rhyming structure of limericks. We report the rhyming distance of the generated limericks, which is calculated as
$$
D_{\text{rhyme}} = \frac{R(l_1, l_2) + R(l_3, l_4) + R(l_1, l_5) + R(l_2, l_5)}{4},
$$
where $l_i$ are the five lines of limericks, and the function $R$ outputs 1 when the two lines rhyme with each other, and 0 otherwise. We report the average distance of 1000 limericks generated by each model respectively, including results from two-stage generation, which uses both forward and reverse models (see Table \ref{rhyming-distance-table}).

\begin{table}
  \caption{Rhyming distance of the generated limericks}
  \label{rhyming-distance-table}
  \centering
  \begin{tabular}{ll}
    \toprule
    Model & Distance $\downarrow$ \\
    \midrule
    GPT-2 (small, forward LM) & 0.9238 \\
    GPT-2 (medium, forward LM) & 0.9598 \\
    GPT-2 (small, reverse LM) & 0.1373 \\
    GPT-2 (medium, reverse LM) & 0.1538 \\
    GPT-2 (small, two-stage) & 0.1810 \\
    GPT-2 (medium, two-stage) & 0.1808 \\
    \bottomrule
  \end{tabular}
\end{table}

\subsubsection{Poetry Quality per Evaluation Metrics}
We report lexical diversity statistics over the original OEDILF limericks dataset, as well as a generated set of free form limericks produced by GPoeT-2 before metric filtering (see Table \ref{lexical-diversity-table}). When filtering generated poems by lexical diversity, we utilize a threshold equal to $\mu_{OEDILF} - 2\times \sigma_{OEDILF}=70.644\%$.



\begin{table}
  \caption{Lexical diversity over input dataset and free form generated poems}
  \label{lexical-diversity-table}
  \centering
  \begin{tabular}{llllll}
    \toprule
    \multicolumn{3}{}{}    &   Lexical Diversity                     \\
    \cmidrule(r){3-6}
    Dataset     & Number of poems   & $\mu$      & $\sigma$  & Max       & Min \\
    \midrule
    OEDILF      & 72,432            & 84.0\%    & 6.678\%     & 100.0\%   & 3.0\% \\
    Free Form   & 12,832               & 77.0\%    & 7.938\%     & 100.0\%   & 16.0\% \\
    \bottomrule
  \end{tabular}
\end{table}

\begin{figure}
     \centering
     \begin{subfigure}[b]{0.49\textwidth}
         \centering
         \includegraphics[width=\textwidth]{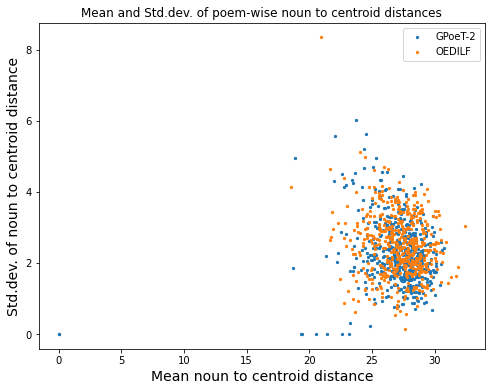}
         \caption{BERT Embedding Distances}
         \label{fig:bert_distances}
     \end{subfigure}
     \hfill
     \begin{subfigure}[b]{0.49\textwidth}
         \centering
         \includegraphics[width=\textwidth]{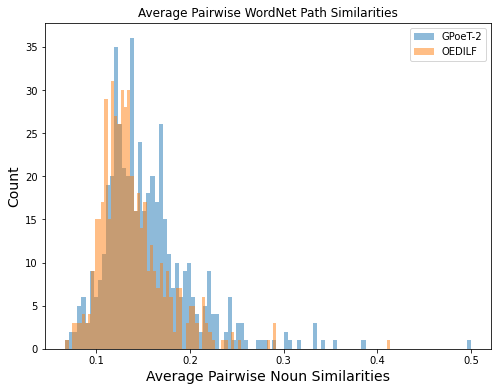}
         \caption{WordNet Distances}
         \label{fig:wordnet_distances}
     \end{subfigure}
     \hfill
        \caption{Metric results on 500 poems each (two-stage model vs. OEDILF)}
        \label{fig:distance_plots}
\end{figure}

Regarding subject continuity metrics, we compared the OEDILF dataset with the GPoeT-2's generated individual poems' noun-to-centroid embedding distances. A plot of these results can be seen in Figure \ref{fig:bert_distances}. Overall, we did not observe any statistical significance in differences between the means of the distributions of the two datasets. While the distributions seem relatively similar in terms of values and density, a few obvious outliers generated by GPoeT-2 have a standard deviation of 0, which is caused by poems only having one or two distinct nouns. Similarly, some poems have a mean distance of 0, indicating that the poem has only a single noun. However, these outliers are relatively rare (around 10 for 500 poems).




We ran a similar experiment for the WordNet subject continuity metric. For each poem we identified the nouns and calculated the path distance for each pair of nouns (using the \texttt{nltk.corpus.path\_similarity(noun1, noun2)} Python function) averaged over the number of pairs. Figure \ref{fig:wordnet_distances} shows the result of this experiment. We did observe statistical significance that the population means of these distributions are likely different (t-test p-value of 2.026-09). Given we believe that this ontology-based metric is a valid proxy for a poem's subject continuity, this implies that GPoeT-2 produces poems indistinguishable from human-written poems with respect to subject continuity.

When running content classification with Google's Content Classification API over all 72,432 OEDILF limericks and over 10,696 lexical diversity filtered free form generated poems, we observe that OEDILF limericks were classified with a success rate (i.e. confidence score greater than 50\% for at least one category) of 52.3\%, while GPoeT-2 poems were classified with a success rate of 51.4\%. Due to the brevity of limericks and prevalence of idiosyncratic words in OEDILF, and GPoeT-2's bias on being fine-tuned on this training set, we presume that this resulted in the lower success rates. We choose to filter out unclassified generated poems as we observed a higher likelihood of better quality generated poems for those that were classified successfully.

\subsection{Qualitative Results}
After filtering and evaluating the generated limericks, we manually selected 94 limericks to form a GPoeT-2 collection. We found that the model was able to generate free form poems from a wide variety of topics without constrained generation. This means GPoeT-2 generates limericks with the \textit{AABBA} rhyming scheme with no seed phrase given and without any prior knowledge or enforcement of the structure. With GPoeT-2, we demonstrate the ability to automatically generate novel, categorized poems across topics like food, nature, and society, providing users with a subset of limericks ready for final selection to be published in an e-book.

Below we present some of these hand-picked limericks produced by GPoeT-2. For more examples, please refer to our collection of 94 categorized limericks in this \href{https://docs.google.com/document/d/1SOekzdPvMrQO9vADkFoHeuIa4zPKk7sBh7CW6hSyxS0/edit#}{e-book}.
\newpage
\begin{multicols}{2}
    An unfortunate fellow named Marge\\
    because he was caught in law's charge\\
    he'd caught out on the law\\
    claimed he hadn't a flaw\\
    his taxes? a number at large\\
    
    This limerick's the last one I'll write\\
    all the ends of five limericks tonight\\
    I'm perplexed through each line\\
    writing limericks that whine\\
    need one limerick? my words are too bright\\
    
    Our society is filled with extremes\\
    we are striving for most of our dreams\\
    we are filled with oration\\
    with the cause of inflation\\
    consumerism: we add to its themes\\
    
    If a tree needs fertilization\\
    to assure the result of creation\\
    this is all that you need\\
    to get free from the seed\\
    then you'll grow for some tree's salvation\\
\end{multicols}

\section{Discussion and Future Work}
We have explored system selection (e.g. fine-tuned GPT-2), variations in language modeling (e.g. standard order, reverse order, two-stage generation), post-processing methods (e.g. grammar correction), and limerick ranking and selection (e.g. lexical diversity). To better measure and further improve the system, we suggest a few potential directions, including
\begin{itemize}
    \item adversarial training or reinforcement learning based on human evaluation result,
    \item custom loss functions stemming from the quality metrics, and 
    \item additional limerick ranking and selection measures.
\end{itemize}

\section{Conclusion}
We present GPoeT-2, an automatic poetry generator that produces reasonable quality limericks that adhere to the \textit{AABBA} rhyming structure. We explore and define "good poetry" metrics, such as grammar correctness, lexical diversity and subject continuity, and utilize them to quantitatively rank, filter, and select generated limericks. Finally, we present a collection of 94 categorized limericks as an e-book, and outline several potential directions for future exploration in this area. 

\small

\bibliographystyle{plainnat}
\bibliography{references}


\end{document}